\documentclass[letterpaper, 10 pt, conference]{ieeeconf}  

\IEEEoverridecommandlockouts                             
\usepackage{graphicx}
\usepackage{caption}
\usepackage{amssymb}
\usepackage{amsmath} 
\usepackage{siunitx}
\usepackage{caption}
\usepackage{subcaption}
\usepackage{array,multirow}
\usepackage{float}
\usepackage{changes}
\usepackage{comment}
\usepackage{amsmath}
\usepackage{xcolor}
\usepackage{multicol}
\usepackage{array,multirow,makecell}
\usepackage{lipsum}
\usepackage{pbox}
\usepackage{breqn}
\usepackage{todonotes}

\title{\LARGE \bf
Neural Network Based Model Predictive Control for an Autonomous Vehicle
}

\author{Maria Luiza Costa Vianna$^{1,2}$, Eric Goubault$^2$ and Sylvie Putot$^2$% <-this % stops a space
\thanks{*This work was supported by the academic Chair "Engineering of Complex Systems", Thal\`es-Dassault Aviation-Naval Group-DGA-Ecole Polytechnique-ENSTA Paris-T\'el\'ecom Paris, and AID project "Drone validation and swarms of drones"}% <-this % stops a space
\thanks{$^{1}$Lab-STICC, ENSTA-Bretagne, Brest, France}%
\thanks{$^{2}$LIX, Ecole Polytechnique, CNRS and Institut Polytechnique de Paris, 91128 Palaiseau, France
        {\tt \{costavianna,goubault,putot\}@lix.polytechnique.fr}}%  
}

\begin{document}

\maketitle
\thispagestyle{empty}
\pagestyle{empty}

%%%%%%%%%%%%%%%%%%%%%%%%%%%%%%%%%%%%%%%%%%%%%%%%%%%%%%%%%%%%%%%%%%%%%%%%%%%%%%%%
\begin{abstract}
 
We study learning based controllers as a replacement for model predictive controllers (MPC) for the control of autonomous vehicles. We concentrate for the experiments on the simple yet representative bicycle model. 
We compare training  by supervised learning  and by reinforcement learning. We also discuss the neural net architectures so as to obtain small nets with the best performances. This work aims at producing controllers that can both be embedded on real-time platforms and amenable to verification by formal methods techniques. 

\end{abstract}

%%%%%%%%%%%%%%%%%%%%%%%%%%%%%%%%%%%%%%%%%%%%%%%%%%%%%%%%%%%%%%%%%%%%%%%%%%%%%%%%
\section{Introduction}

In the past few decades, neural networks have gained in popularity in control and in particular in  optimal control~\cite{bertsekas2019reinforcement}. 
We can distinguish two main approaches for training neural networks in the context of control: supervised and reinforcement learning (RL).

In supervised learning, labeled data are used. The neural network learns to mimic the behavior of a controller, mapping the inputs to known outputs, called labels, that can for instance be generated by a classical controller.
One of the first works in the field of autonomous vehicles controlled by a neural network was proposed in \cite{alvinn}. Using a small feed-forward neural network with only one hidden layer, the network outputs a discrete steering angle value. The training data set was based on the choices of steering angle of a human driver. The author concluded that even though more consistent, the human driver had an inaccurate estimate of the center of the road whereas neural network controllers could drive the vehicle closer to the road center. 

In RL, the system follows a trial and error method by interacting with its environment. In e.g. \cite{Rajeev},  RL is used to derive a controller for the steering angle of an autonomous vehicle from Lidar measurements. The authors also study the safety of the  system, concluding that the neural net controller is not robust to lidar faults, and that the high-dimensionality of the problem is not easing the verification.

In this work, we train neural networks both by supervised  and reinforcement learning, as a replacement for a Model Predictive Controller (MPC, \cite{Richalet}) of an autonomous vehicle.
At each time-step, an MPC receives the current state of the system and optimizes the set of consecutive control commands that must be applied in order to reduce  the error between the propagated system's state and a reference state, over a finite time horizon. The problem is formulated with a cost function that is minimized under the constraints and the dynamic equations of the system. 
MPC offers high performance controllers but the optimization process is computationally expensive, which restricts its use for embedded systems. 
Some solutions have been proposed: explicit MPC \cite{bemporad}, for instance, pre-computes an offline  solution, allowing a fast evaluation of the control problem online.  
However, this approach assumes that the controlled system is linear and time-invariant. In our context, the choice to use neural networks is motivated by the fact that they are general function approximators that can account for non-linearity.
 
Representing MPC controllers with a neural network trained by a supervised learning approach is a well-studied problem \cite{ortega,Bernt,Parisini,Spielberg,Georges,Lucia}. 
Our objective in this work is to further investigate how  the different hyper parameters, the inputs of the neural network and the complexity of the training data set, affect its capacity to learn the MPC's behavior. % 

The  goal is to choose the smallest and simplest architecture of the network in order to make the evaluation of the neural net faster, and to facilitate the process of verification of the safety of the controller, as in e.g. \cite{sherlock,Dutta,verisig}. 
Indeed, the size of the network controller increases the computational complexity of verification and is even more determinant when the verification is done online, as in e.g. \cite{online5,online3}.

\subsection*{Contributions and Contents}

- We study the behaviour of the trained neural network depending on different input sets representing data sets of future reference points on trajectories; 

- We analyze the effect of varying the number of layers on the performance  of the neural controller;

- We analyze the effect of varying the activation functions for the hidden layers; %

- We demonstrate that our selected network architecture is capable of generalizing to a real world scenario after being trained with data sets constructed over basic trajectories. In addition, we demonstrate that a very small neural network can learn from the MPC expert with great precision;
 
- Finally, we compare the performance of supervised learning with RL for representing an MPC policy.

We demonstrate that both learning methods are able to represent the MPC problem accurately  while being much more efficient in terms of computational complexity than the original algorithm.  

The remainder of this paper is organized as  follows. In Section \ref{setting}, we introduce the model that motivates this work. The supervised learning approach is presented in Section \ref{suplearning} and the reinforcement learning approach in Section \ref{rlearning}. Experimental results are discussed in Section \ref{experiments}. %

\subsection*{Related work}

In \cite{Georges,Lucia,Lucia2} authors use the relative system's state with respect to the reference trajectory as input set. %

In \cite{Bernt}, the input set is composed by the system's state along with information about previous control decisions and the reference for the immediate next system's output. This is a classical view in control, but not in optimal or model-predictive control, where we seek an optimal controller with respect to a bounded time horizon. %
In \cite{ortega}, a parametrization of the reference trajectory is adopted to reduce the number of inputs. However, we obtained better experimental results when adopting a more straightforward approach, using input sets that are closer to the input set received by the MPC, thus avoiding approximating the reference trajectory.   

The literature is divided about the benefits of using shallow or deep networks. In \cite{Georges}, the author did not see significant improvement when increasing the number of layers on the neural controller. In contrast, in \cite{Lucia2},  the authors experienced an improvement of the controller using a deep network over a one layer network when learning a non-linear MPC.

When approximating a linear MPC with quadratic objective (LQR) by a neural network, it is natural to use ReLU as activation function~\cite{Lucia} since the exact solution to the corresponding MPC problem is piecewise linear.
The situation is however less obvious for  non-linear problems. 

In \cite{LinNasser}, the authors compare the performance of the classical MPC algorithm with a neural controller trained with  the Deep Deterministic Policy Gradient (DDPG) algorithm \cite{ddpg}. They concluded that, when the dynamics of the system are known, the  RL approach has a performance close to the one obtained by the MPC. However, and this is beyond the scope of our article, they observed that, when modeling errors are included, RL is significantly better than MPC.

\section{Vehicle's model} \label{setting}
 
We consider an autonomous vehicle equipped with a lidar, which can be used to compute a reference trajectory under the form of waypoints. 
We choose the simple bicycle model with constant speed $v_{w}$ as the dynamic of the vehicle, %We will also simplify further the model so that the speed is constant. 
where the controller acts  on the steering angle only, given by 
\begin{equation}
	\label{eq::propagate}
	X^{k+1}_{w} =  f(X^{k}_{w},{\delta}^k) =  \begin{bmatrix}
	x^k_{w} + v_{w}\cdot cos(\theta^k_{w})\cdot dt \\
	y^k_{w} + v_{w}\cdot sin(\theta^k_{w}) \cdot dt \\
	\theta^k_{w} + \frac{v_{w}}{l_f} \cdot sin({\delta}^k) \cdot dt
	\end{bmatrix}
\end{equation}

\begin{equation}
	\label{eq::observe}
	Y^{k}_{w} =  h(X^{k}_{w}) =  \begin{bmatrix}
	1 & 0 & 0 \\
	0 & 1 & 0
	\end{bmatrix}\cdot X^{k}_{w}
\end{equation}

where $$X_{w}=  \begin{bmatrix}
x_{w} & y_{w} & \theta_{w}
\end{bmatrix}^T, \; \;
%$$
%$$
Y_{w} =  \begin{bmatrix}
x_{w} & y_{w}
\end{bmatrix}^T 
$$

We use  uppercase letters to represent vectors and matrix, and note $X_{w}$ the system's state and $Y_{w}$ its output, observable at each time step. Lowercase letters represent scalars, $x_{w}$ and $y_{w}$ are the 2-dimensional representation of the vehicle's position, $\theta_{w}$ its orientation and $v_{w}$ its longitudinal speed with respect to the frame $w$. Subscripts indicate the reference frame in which the value is represented: the letter $w$ stands for world, representing a global fixed frame. The letter $r$ stands for robot and it represents the vehicle's frame. 
In Equation (\ref{eq::propagate}), $l_f$ is a constant that represents the distance from the vehicle's front to its gravity center, and the steering angle ${\delta}^k$ is the control input. 
Superscripts indicate the time step. Therefore, $X_{w}^0$ is the initial state vector of the vehicle represented in the global frame. 
Finally, we use the symbol \~{} to indicate a reference value. For example, $\tilde{Y}_{w}$ is the expected system's output. In the problem addressed in this work the reference is defined by points on the trajectory the vehicle is supposed to follow.

\section{Data set}
\label{data set}
In this section, the MPC equations are defined and then used to create the training and the validation data sets. 
\subsection{The Expert}

An MPC problem consists in computing the vector $U= \begin{bmatrix}
u_{k} & \hdots & u_{k+N-1} 
\end{bmatrix}$, where $u_k \in {\Bbb R}^m$ is the control input  of the system at step $k$, that minimizes the error between ${Y_{w}}$ and $\tilde{Y}_{w}$ over the $N$ next time steps, $N$ being the control horizon:
\begin{dmath}
	J(U) =  \sum_{i = k + 1}^{k+N}  (Y_{w}^i - {\tilde{Y}^i_{w}})^2
	\label{costmpc}
\end{dmath} 
with $X_{w}^{k+1} = f(X_{w}^{k},u_k)$ given by Equation (\ref{eq::propagate}) and $Y_{w}^{k} = h(X_{w}^{k})$ given by Equation (\ref{eq::observe}).

At each time step $k$, $k =0,1,\hdots,T$, 
the first control value $U[0] = u_{k}$ is used as input, the other values of the vector are discarded. 
The MPC relies for the optimization on the input vector $$I_{mpc}^k =  \begin{bmatrix}
X_{w}^{k} & {\tilde{Y}^{k+1}_{w}} & \hdots & {\tilde{Y}^{k+N}_{w}} \end{bmatrix}.$$

We  consider $u_k= 
{\delta}^k$ 
with a control horizon of 20 steps. This value was defined empirically by considering the trade-off between a control horizon that if too big would minimize the importance of close steps and if too small would not be able to react in time to changes on the path. 

\subsection{Training Data set}

The training data set is composed of input sets along with labels that correspond to the expected outputs, generated using the MPC algorithm. The labels are used only by the supervised learning method.
 
We create three  training data sets corresponding to sample trajectories of increasing complexity. Each data set has the same number of samples.
 
For each sample, the relative initial position of the vehicle with respect to the trajectories is chosen randomly. 
Data set 1 contains only straight trajectories. Data set 2 has straight trajectories and two sinusoidal trajectories, with one frequency being twice the other. These different frequencies allow us to simulate smooth and abrupt curves, so that the network can adapt to different turning rates. Data set 3 has spiral trajectories in addition to trajectories of Data set 2.

\subsection{Validation Data set}
\begin{figure}[!h]
	\centering
	\includegraphics[height=2.5cm,width=6cm]{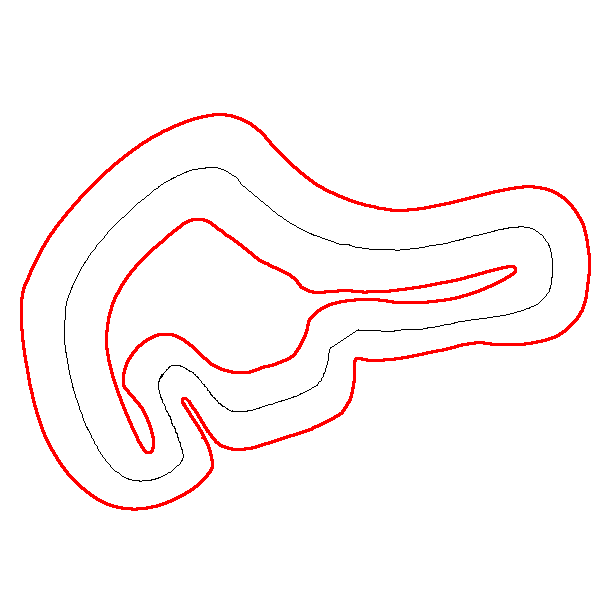}
	
	\caption{Validation data set: reference trajectory}
	\label{track-img}
\end{figure}

The validation data set is completely independent of the training data set in order to prove that the trained network is capable to generalize to new information. We consider the F1tenth \cite{f1tenth} race track from the Berlin 2020 virtual competition  
on which we define the reference waypoints so as to follow a centerline strategy. 
We compare the different networks for two laps on the race track, Figure \ref{track-img}.

\vspace{-0.5cm}

\section{Supervised Learning}\label{suplearning}
 
The range of values returned by the networks is bounded due to physical constraints on the vehicle. We choose to use the hyperbolic tangent function as the  activation function for the last layer, so as to smoothly enforce these bounds. 
We use as loss function the mean squared error between the steering angle predicted by the network and the one calculated by an MPC controller,

\begin{equation}
\label{costbackp}
J= \frac{1}{M}\sum_{m = 0}^{M} (\delta_{m}  - \tilde{\delta}_{m} )^2
\end{equation}
where $M$ is the number of samples in the training data set.

\subsection{Network's Input Set}\label{inputset}
We first study the impact of the choice of the input set, in particular the amount of future points on the reference trajectory given as input to the network. Information is not interpreted and treated the same way by the MPC and by the neural network.  While knowing the system's state in a global reference frame is important for the MPC because it solves the system's differential equations, it is not necessary for the neural network based controller. We thus take as input of the neural controller the reference trajectory expressed in the vehicle's reference frame instead of in the global frame as in $I_{mpc}$. This removes the need for encoding the vehicle's state in addition to the reference points as inputs to the network. Anyway, the controls should not depend on the global position of the system, but rather on the relative configuration of the system with respect to the target trajectory.

The input set ${I^k_{3}}$ contains only the closest reference point to the vehicle's position, ${\tilde{Y}^k}$.  In addition, it contains the angle ${\theta}^k$ that represents the expected orientation of the vehicle at the reference point ${\tilde{Y}^k}$ which expresses some knowledge about future reference points:
\begin{equation}
\label{i3}
	{I^k_{3}} =  \begin{bmatrix}
	{\tilde{Y}^k_r} & {\theta}_r^k
	\end{bmatrix}
\end{equation}

The  second input set ${I^k_{N+1}}$ includes only the $y$ component of the N reference points ${\tilde{Y}_r}$, plus the orientation ${\theta}^k_r$:
\begin{equation}
\label{i21}
{I^k_{N+1}} = {I^k_{21}} = \begin{bmatrix}
{\tilde{y}_r^{k+1}} & \hdots & {\tilde{y}_r^{k+N}}  & {\theta}^k_r
\end{bmatrix}
\end{equation}

The third set ${I^k_{2N}}$  is composed of the position of the $N$ reference points used by the  MPC at each time step:  
\begin{equation}
\label{i40}
{I^k_{2N}} = {I^k_{40}} =\begin{bmatrix}
{\tilde{Y}^k_r} & \hdots & {\tilde{Y}_r^{k+N}}
\end{bmatrix}
\end{equation}

\subsection{Network architecture}
A fully connected neural network is defined as:
\begin{eqnarray} 
O_{1} & = & a_{1}(W_1 I_{n}  + B_1),\nonumber\\
\vdots \nonumber \\
O_{n_h + 1} & = & a_{n_h + 1}(W_{n_h + 1} O_{n_h}  + B_{n_h + 1})
\label{outnn}
\end{eqnarray}
where $n_h $ is the number of hidden layers, $I_{n} $ is the input, and $W_i$, $B_i$, $a_i$ and $O_i$ are respectively the weights, biases, activation function and output of the $i_{th}$ layer, $i = 1,\hdots,n_h + 1$.

We will vary the network hyperparameters: the number of hidden layers and the number of neurons in each layer, and will analyze their impact on the performance of the neural network.
Each hidden layer adds a step of calculation between the input and the output of the network, and together with the number of neurons within each layer defines the complexity of the network evaluation.

Dividing a fixed amount of neurons through layers increases the number of computation steps. Moreover, as far as formal verification is concerned, higher number of layers usually means more conservative results for the more scalable abstraction-based techniques. We keep this trade-off in mind while analyzing the performance of the neural controllers.

\section{Reinforcement Learning}
\label{rlearning}
We use here the DDPG algorithm \cite{ddpg}.  
This approach considers a continuous state space and output space and therefore is well suited for  control problems \cite{Lin,Yu}. 
The DDPG algorithm is an actor critic algorithm. In a nutshell, this means that two neural networks are trained during the process. The actor predicts the action that should be applied to the system given the system's state. The critic receives as input the system's state and the action predicted by the actor and analyzes the actor's choice relying on a certain value function. General RL algorithms try to maximize value functions of the form:
\begin{equation}
\label{real_critic}
Q(s_t,a_t) = r_{t} + \gamma r_{t+1} + \gamma^2 r_{t+2} + \hdots + \gamma^{N}r_{t+N} + \hdots
\end{equation}
\noindent where we take $N$ as the time horizon of the MPC, $\gamma$ is the discount factor (between 0 and 1) and $r_t$ is the immediate reward at time $t$ that we take as: 
\begin{equation}
\label{ideal_reward}
r_{t}  = - (Y_{w}^{t+1} - {\tilde{Y}^{t+1}_{w}})^2.
\end{equation}
The DDPG algorithm maximizes this $Q$-function represented by the critic network for all states described in a pre-defined environment, by sampling this state space over a possibly large set of training episodes. The actor network is the one memorizing the best actions to take for reaching the maximal reward. Given the current state of the system is , i.e., when training has converged, the end neural net controller that we expect to use in lieu of the MPC.

The $Q$-function of Equation (\ref{real_critic}) is close to being the opposite of the objective function to minimize in the MPC formulation for $\gamma=1$, making the RL task for the reward defined by Equation (\ref{ideal_reward}) ideally suited for replacing an MPC. Still, there are two major differences. The first one is that we used continuous training so the $Q$-function is defined on a potentially infinite time horizon. 
The second difference is that to maintain convergence stability, the biggest value we can adopt for $\gamma$  is $0.99$, see \cite{lavet}. This actually counter-effects the first difference we mentionned: with a discount factor of 0.99, the effect of the distance to the reference trajectory after more than 250 steps becomes in fact quite negligible with respect to the first $N=20$ steps. %

We chose to use as environment, for representing the system's state space in the DDPG algorithm $I_{rl}=I_{40}$, even though the reward function needs only the first reference point in $I_{40}$. This makes the information of the future mismatch between the actual trajectory and the reference trajectory known to DDPG, over the $N$ steps time horizon. 

\section{Tests and Results}  \label{experiments}
We now present and discuss the results of our experiments, first on the choice of the best architecture in the supervised learning approach, then compared to reinforcement learning. 
Experiments are performed on an Intel Core i7 running Ubuntu 18.04.3 at 1.9 GHz. We use the Open AI Gym 0.17.2,  
TensorFlow 1.5.0 and Keras 2.1.5 frameworks for RL and supervised learning tasks. Learning rate is set to 0.001 and in the case of DDPG, we use replay buffers of size 2000. 
The system's parameters are set to $v_w = 3.0$ $m/s$ and $l_f = 0.15875$ $m$.

\subsection{Measurement of performance}
In this section, we compare different networks, evaluating the performance of the neural network to learn from the expert to follow a reference trajectory. 

The network's  performance can be quantified by the mean squared error on the validation data set, Equation (\ref{costbackp}).
However, a more representative measure of the capacity to follow a reference trajectory is the error  between the expected and the realized trajectories. We thus mostly focus on the mean error, maximum error and standard deviation between the neural network controlled trajectory
and the trajectory realized by the vehicle controlled by the classical MPC controller when following the reference trajectory.

\subsection{Supervised Learning}
\subsubsection{Input set}
\begin{figure*}[!t]
	\centering
	\includegraphics[scale=0.44]{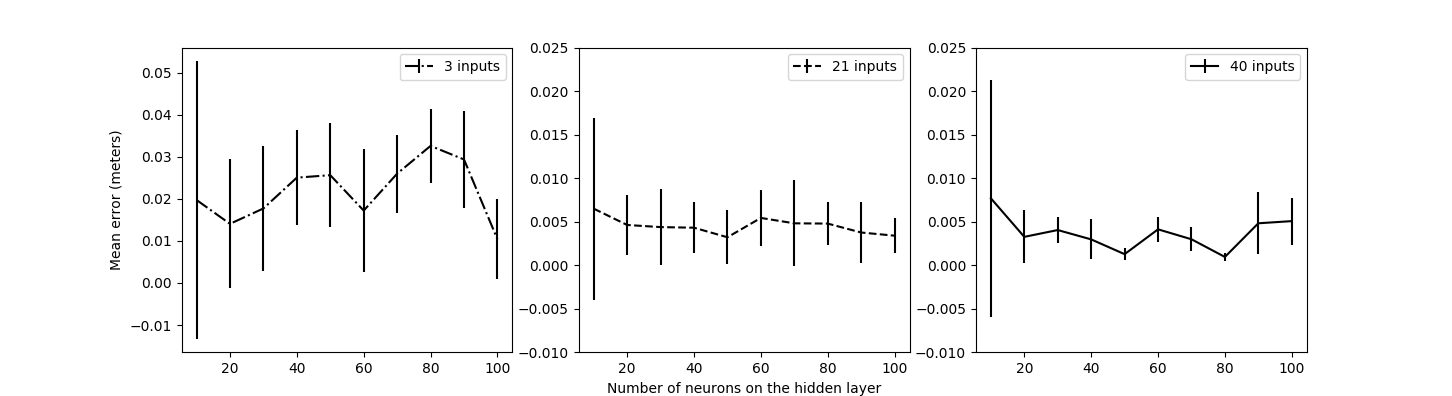}
	\caption{Choice of the input set: mean error and standard deviation between the MPC trajectory and the neural network controlled trajectory on the validation data set.}
	\label{analysis1-2}  
\end{figure*}

\begin{figure}[h]
	\centering
	
	\includegraphics[scale = 0.44]{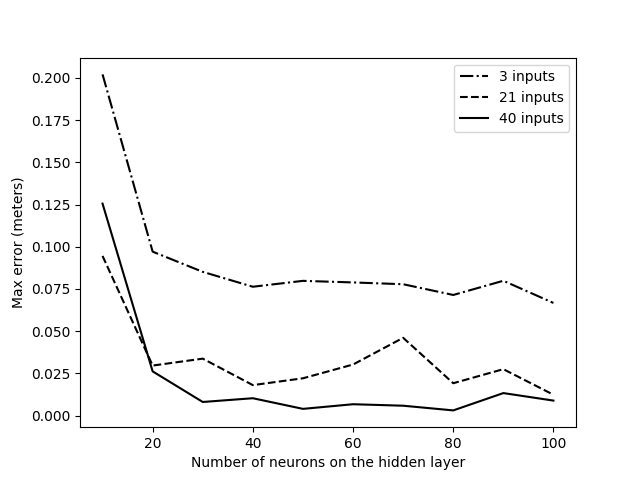}
	\caption{Choice of the input set: maximum error}
	\label{analysis1-1} 
\end{figure}

We first analyze in Figures \ref{analysis1-2} and  \ref{analysis1-1} the impact of the choice of the different input sets defined in Section \ref{inputset} on the results given on the validation data set. This is done for a shallow network with one hidden layer, in which we vary the number of neurons. 
 
Too many neurons per layer may cause an overfitting of the training data set which makes generalization to new inputs difficult. On the other hand, with too few neurons the network may not be able to learn a pattern and represent the complexity of the expert function.  

Looking at Figures \ref{analysis1-2} and  \ref{analysis1-1}, we do not see an important trend to overfit at least till 100 neurons.
Indeed, in terms of maximum error (Figure \ref{analysis1-1}), we note that the best network structure for the three input set configurations needs a considerably large amount of neurons per layer, 100 for $I_{3}$ and $I_{21}$ and 80 for $I_{40}$. 
 However, rather than trying to increase the size to improve performance, we focus in what follows on improving the performance of smaller networks.

We note that the performance of the networks in terms of errors are significantly better for input sets $I_{21}$ and even more for $I_{40}$ compared to the smaller input set $I_{3}$.
 
Moreover, networks with $I_{40}$ have smaller standard deviation values and therefore are more robust along the reference trajectory. This result is intuitive since $I_{40}$ is an input set closer to ${I_{mpc}}$ which facilitates the task of learning from the MPC.
Still, all studied inputs were able to represent the MPC problem reasonably accurately. Depending of the problem, $I_{3}$ might be considered precise enough. The advantage of this input set is that it considerably reduces  the size of the neural network and has a fixed size whatever the control horizon.

\subsubsection{Number of hidden layers}
We now fix the input set to $I_{40}$ and vary the number of hidden layers from one to three and keep varying the number of neurons per layer.
 
\begin{figure*}[!t]
	\centering
	
	\includegraphics[scale = 0.44]{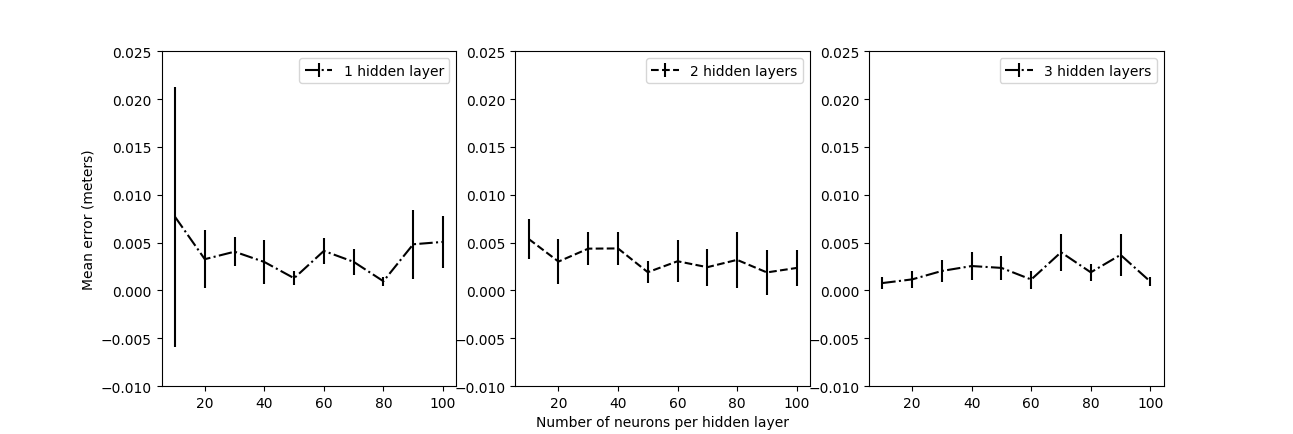}
	\caption{Comparing shallow to deep networks: mean error  and standard deviation between MPC and neural network trajectory}
	\label{analysis2-1}  
	\vspace{-0.5cm}
\end{figure*}

\begin{figure}[!h]
	\centering
	
	\includegraphics[scale = 0.44]{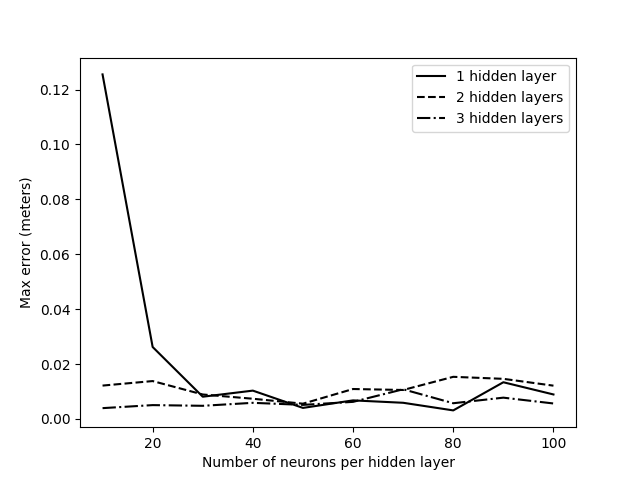}
	\caption{Comparing shallow to deep networks: maximum error}
	\label{analysis2-2} 
\end{figure}

Let us consider the maximum and mean error represented in Fig. \ref{analysis2-1} and \ref{analysis2-2}. We observe that networks with 2 and 3 hidden layers perform better than 1 hidden layer for smaller amounts of neurons. As the number of neurons increase, however, shallow networks tend to be more accurate. This pattern may be the result of an overfit on the training data set.

Increasing the number of hidden layers allows us to use a smaller amount of neurons per layer and even a smaller total number of neurons. 
We can observe on Figures \ref{analysis2-1} and \ref{analysis2-2}  that three hidden layers  allow us to have good results with only 10 neurons per layer, while one hidden layer would require around 80 neurons per layer for a similar precision. Their evaluation in terms of running time is comparable (6.75$\si{\micro}$s for the 1 layer case for 5.84$\si{\micro}$s for the 3 layers case, to be compared to the 6.75 ms necessary for running the MPC). 
This makes the 3 hidden layer structure a good trade-off in terms of complexity with respect to performance.

Based on these considerations, we now use networks with 3 hidden layers and 10 neurons per layer in what follows.

\subsubsection{Training data set}
Until now, we were using the complete Data set 3. We now compare in Table \ref{tab4} the results of the networks trained with Data sets 1, 2 and 3. 
\begin{center}
	\captionof{table}{Impact of the training data set}
	\label{tab4}
	\begin{tabular}{|c|c|c|c|}
		\hline \vtop{\hbox{ \strut \textbf{Data sets }}}  & 1  & 2 &  3 \\
		\hline \vtop{\hbox{\strut \textbf{Maximum error (cm)}}} & 4.6  & 1.3  & 0.39  \\ 
		\hline \vtop{\hbox{\strut \textbf{Mean error (cm)}}} & 1.88 & 0.14  & 0.077 \\ 
		\hline \vtop{\hbox{\strut \textbf{Standard deviation (cm)}}}  & 0.96  & 1.23  & 0.061  \\ 
		\hline
	\end{tabular}
\end{center}

As expected, the neural controller's performance improves on the validation data set when the training data set covers more types of trajectories. However, it is interesting to analyze that learning from a data set where very simple trajectories such as straight lines are followed, the network is still capable to generalize it with reasonable precision in more complex contexts. This can be explained by the fact that at each time step, both the MPC and the neural net  controller receive only a small portion of the whole reference trajectory, only 20 steps ahead. In this case, the trajectory might be momentarily close to a linear trajectory. 

\subsubsection{Activation function}
All previous analyses were done using ReLU as activation function in the hidden layers. Now, we vary the activation function between tanh, sigmoid and ReLU in the hidden layers. In the output layer, we keep using tanh.
Table \ref{tab3} presents the results: we observe a clear improvement on the precision when using the sigmoid.
\begin{center}
\captionof{table}{Impact of the activation function}
	\label{tab3}
	\begin{tabular}{|c|c|c|c|}
		\hline \vtop{\hbox{ \strut \textbf{Activation function}}}  & ReLU  & tanh &  sigmoid \\
		\hline \vtop{\hbox{\strut \textbf{Maximum error (cm)}}}  & 0.39 & 0.42 & 0.17 \\ 
		\hline \vtop{\hbox{\strut \textbf{Mean error (cm)}}}  & 0.077 & 0.126 & 0.059 \\ 
		\hline \vtop{\hbox{\strut \textbf{Standard deviation (cm)}}}  & 0.061 & 0.091 & 0.03 \\ 
		\hline
	\end{tabular}
\end{center}

\subsection{Reinforcement Learning}
We were not able to obtain a precise enough controller unless the size of the actor network was considerably increased. We use the architecture proposed in \cite{ddpg}, a two layers neural network with 400 and 300 neurons respectively, to obtain the results presented in Table \ref{tab5} on the same validation data set as for supervised learning. 
 
During validation, the neural controller receives  as reference the same trajectory of a vehicle controlled by the classical MPC as received in the supervised learning approach.

\begin{center}
	\captionof{table}{Results for the reinforcement learning approach.}
	\label{tab5}
	\begin{tabular}{|c|c|}
		\hline \vtop{\hbox{\strut \textbf{Maximum error (cm)}}}  & 0.97 \\ 
		\hline \vtop{\hbox{\strut \textbf{Mean error (cm)}}}  & 0.49 \\ 
		\hline \vtop{\hbox{\strut \textbf{Standard deviation (cm)}}}  & 0.16 \\ 
		\hline
	\end{tabular}
\end{center}

 Table \ref{tab5} shows that the neural controller trained by RL also represents the MPC problem with good precision, although we could not reach the same precision as with supervised learning. 
The RL approach presents the disadvantage of the networks dimensions when compared with the neural network's architecture with supervised learning. To fix ideas, the evaluation of the RL network takes 0.11ms per step, to be compared with the approximately 6$\si{\micro}$s with the supervised learning approach and the 6.75ms with the plain MPC. On the other hand, it presents the advantage of not needing to explicitly know the expert.

\section{Conclusion}\label{conclusion} 
We trained neural networks  with supervised and reinforcement learning to replace MPC controllers in autonomous vehicles. 
For supervised learning, we analyzed how different neural architectures affect the process of learning. 
We varied the input set, and confirmed that providing all future points in the control horizon to the neural network facilitates the learning process.
We verified that our neural network model is capable of generalizing to real problems while learning with a data set constructed on a few simple trajectories. This is possible because the input set used for the neural network takes into consideration the relative position of the vehicle with respect to the reference trajectory and because the MPC considers only a small portion of the reference trajectory at each time step. 
We observed that for supervised learning, sigmoid activation functions in the hidden layers were best suited than the classical alternatives. 
Adding hidden layers in a neural network is more expensive, in terms of computational cost, than increasing the total number of neurons on the network. We showed that to represent our non-linear MPC problem, 3 hidden layers allowed to obtain networks with excellent performance with only 10 neurons per layer, providing a good trade-off between precision and cost.

The  networks obtained  with  supervised and reinforcement learning were both able to successfully approximate a classic MPC with a maximum deviation of less than 1 centimeter from the trajectory controlled by MPC. The neural approximator is thus able to replace the classical controller while being much more computationally efficient.

Finally, reinforcement learning was found to be less performing than  supervised learning in our context where the dynamic of the system is well known. This might not be the case, however, when modeling errors are included. 

For future work, we intend to focus on the formal verification of safety properties on the neural network controllers in the line of e.g. \cite{verisig,Dutta}, and on some criteria for the selection of neural networks that make them amenable to verification, beyond the relatively obvious criteria such as dimension.  
%%%%%%%%%%%%%%%%%%%%%%%%%%%%%%%%%%%%%%%%%%%%%%%%%%%%%%%%%%%%%%%%%%%%%%%%%%%%%%%%

%%%%%%%%%%%%%%%%%%%%%%%%%%%%%%%%%%%%%%%%%%%%%%%%%%%%%%%%%%%%%%%%%%%%%%%%%%%%%%%%

%%%%%%%%%%%%%%%%%%%%%%%%%%%%%%%%%%%%%%%%%%%%%%%%%%%%%%%%%%%%%%%%%%%%%%%%%%%%%%%%
%\section*{APPENDIX}


%\section*{ACKNOWLEDGMENT}

%%%%%%%%%%%%%%%%%%%%%%%%%%%%%%%%%%%%%%%%%%%%%%%%%%%%%%%%%%%%%%%%%%%%%%%%%%%%%%%%

\end{document}